\documentclass[final]{cvpr}

\usepackage{times}
\usepackage{epsfig}
\usepackage{graphicx}
\usepackage{amsmath}
\usepackage{amssymb}

\usepackage[pagebackref=true,breaklinks=true,colorlinks,bookmarks=false]{hyperref}

\usepackage{floatrow}
\usepackage{graphicx}
\usepackage{floatrow}
\usepackage{comment}
\usepackage{amsmath,amssymb} 
\usepackage{color}
\newfloatcommand{figurebox}{figure}[\nocapbeside][\dimexpr(\textwidth-\columnsep)/2\relax]
\newfloatcommand{tablebox}{table}[\nocapbeside][\dimexpr(\textwidth-\columnsep)/2\relax]

\usepackage{array}
\newcolumntype{C}[1]{>{\centering\arraybackslash}m{#1}}
\newcolumntype{R}[1]{>{\raggedleft\arraybackslash}m{#1}}
\newcolumntype{P}[1]{>{\raggedright\arraybackslash}p{#1}}
\newcolumntype{M}[1]{>{\centering\arraybackslash}m{#1}}
\usepackage{multirow}
\usepackage{algorithm}
\usepackage{algorithmic}

\floatstyle{ruled}
\restylefloat{algorithm}
\usepackage[font=small,labelfont=bf,labelsep=period]{caption}
\usepackage{xcolor}

\def\cvprPaperID{1821} 
\def\confYear{CVPR 2021}

\begin{document}

\title{Group-aware Label Transfer for Domain Adaptive Person Re-identification}

\author{ Kecheng Zheng$^{1}$\thanks{Equal contribution}, Wu Liu$^{2*}$, Lingxiao He$^{2}$, Tao Mei$^{2}$, Jiebo Luo$^{3}$, Zheng-Jun Zha$^{1}$\thanks{Corresponding author}\\
\\
{\small $^1$University of Science and Technology of China, $^2$AI Research of JD, $^3$University of Rochester}\\
{\small zkcys001@mail.ustc.edu.cn,\ \{liuwu1,helingxiao3,tmei\}@jd.com,\ jluo@cs.rochester.edu,\ zhazj@ustc.edu.cn}}

\maketitle

\begin{abstract}

Unsupervised Domain Adaptive (UDA) person re-identification (ReID) aims at adapting the model trained on a labeled source-domain dataset to a target-domain dataset without any further annotations. Most successful UDA-ReID approaches combine clustering-based pseudo-label prediction with representation learning and perform the two steps in an alternating fashion. However, offline interaction between these two steps may allow noisy pseudo labels to substantially hinder the capability of the model.
In this paper, we propose a Group-aware Label Transfer (GLT) algorithm, which enables the online interaction and mutual promotion of pseudo-label prediction and representation learning. 
Specifically, a label transfer algorithm simultaneously uses pseudo labels to train the data while refining the pseudo labels as an online clustering algorithm. It treats the online label refinery problem as an optimal transport problem, which explores the minimum cost for assigning M samples to N pseudo labels.
More importantly, we introduce a group-aware strategy to assign implicit attribute group IDs to samples. The combination of the online label refining algorithm and the group-aware strategy can better correct the noisy pseudo label in an online fashion and narrow down the search space of the target identity. 
The effectiveness of the proposed GLT is demonstrated by the experimental results (Rank-1 accuracy) for Market1501$\to$DukeMTMC (82.0\%) and DukeMTMC$\to$Market1501 (92.2\%), remarkably closing the gap between unsupervised and supervised performance on person re-identification.~\footnote{Full codes are available in https://github.com/zkcys001/UDAStrongBaseline and https://github.com/JDAI-CV/fast-reid}
\end{abstract}

\section{Introduction}

	\begin{figure}[th]
		\centering
		\includegraphics[width=1.0\linewidth]{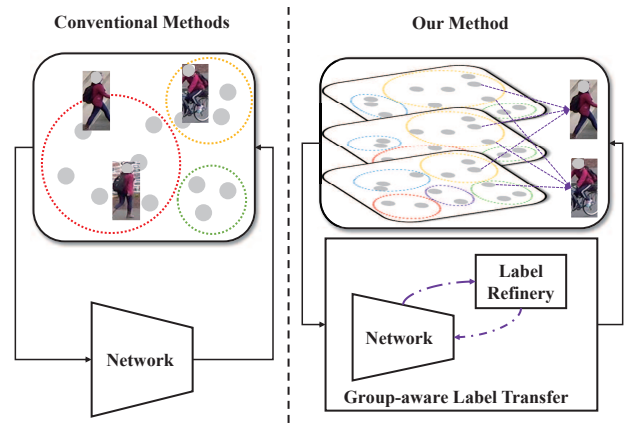}
		\caption{Illustration of conventional methods and our group-aware label transfer method. 
		In our method, each instance is assigned to multiple prototypes with different granularity for generating multi-group pseudo labels, and then its noisy multi-group pseudo labels are online refined. 
		By learning the refining multi-group pseudo labels, our GLT method can learn an embedding space that encodes the semantic multi-granularity structure of data.}
		\label{fig:intro}
	\end{figure}

Person re-identification (ReID) is the important task of matching person images captured from non-overlapping camera networks, which is widely used in practical applications such as automatic surveillance, content-based retrieval, and behavior analysis~\cite{liao2015person,yang2014salient,zheng2011person,dai2021CVPR}. It has been proved that existing approaches can achieve remarkable performance when the training and testing data are collected from the same application scenario but often fail to generalize well to other scenarios due to the inevitable domain gaps. Therefore, it is necessary for both academia and industry to study the Unsupervised Domain Adaptive (UDA) person re-identification (ReID) problem.


Existing UDA-ReID approaches~\cite{ge2020mutual,song2018unsupervised,zhai2020adcluster,zhong2019invariance} typically include three steps: feature pre-training with labeled source domain data, clustering-based pseudo-label prediction for the target domain data, and feature representation learning/fine-tuning with the pseudo-labels. The last two steps are usually iteratively conducted to strengthen or promote each other. However, the first problem is that the pseudo-labels assigned through clustering usually contain incorrect labels due to the divergence/domain gap between the source and target data, and the imperfect nature of the clustering algorithm. 
Such noisy labels may mislead the feature learning and harm the domain adaptation performance. Although the label refining objective in clustering is tractable, it is an offline time-consuming scheme as it requires a pass over the entire dataset. Therefore, online refining those incorrect samples when training can help model learn more robust and accurate representation.

Another problem is that the target domain lacks ID information, so it is difficult to cluster the person images according to human identity. However, in the real world, each person has their own characteristics in his or her appearance. There may be common appearances shared by a group of people but they are not the same identity (e.g. two men with the same red coats and similar black pants as shown in Fig.~\ref{fig:intro}). Therefore, group-based description~\cite{kim2020groupface} that involves common characteristics in a pseudo group, can be useful to narrow down the set of candidates and beneficial to identify the exact persons. This group-aware strategy can cluster a person into multi-group clustering prototypes, and is able to efficiently embed a significant number of people and briefly describe an unknown person. Inspired by this, as shown in Fig.~\ref{fig:intro}, combining the online label refining algorithm with the group-aware strategy may be beneficial to the success of domain adaptation.

In this paper, we propose a Group-aware Label Transfer (GLT) algorithm that facilitates the online interaction and mutual promotion of the pseudo labels prediction and feature learning. 
Specifically, our method simultaneously uses pseudo labels to train the data while online refining the pseudo labels via the label transfer algorithm.
This label transfer algorithm regards the resulting label refining problem as optimal transport, which explores the minimum cost for assigning M samples to N pseudo labels. This problem can therefore be solved by the Sinkhorn-Knopp algorithm~\cite{cuturi2013sinkhorn} of linear programming in polynomial time.
Meanwhile, the label transfer algorithm is scalable, which can be trained with several batches or the whole dataset and can scale to unlimited amounts of data. 
More importantly, we introduce a group-aware strategy to assign implicit attribute group IDs to samples. Explicit grouping requires manual annotation. Therefore, we adopt a group-aware clustering algorithm to generate multi-group pseudo labels. By adopting the concept of grouping, the ReID network can reduce the search space and flexibly embed a significant number of identities into an embedding feature. As shown in Fig.~\ref{fig:intro}, we combine the online label refining algorithm with the multi-group pseudo labels. The combination of the online label refining algorithm and the group-aware strategy can better correct the noisy pseudo label online and narrow down the search space of the target identity and predict more accurate pseudo labels.
In addition, we design a target instance memory bank combined with weighted contrastive loss~\cite{sun2020circle} to make the model more powerful for features representation. 
The proposed GLT framework achieves state-of-the-art performance on Market-to-Duke, Duke-to-Market, Market-to-MSMT, and Duke-to-MSMT with unsupervised domain adaptation, and in particular the \ie Duke-to-Market performances (92.2 Rank-1 and 79.5 mAP) are almost comparable with the supervised learning performances (94.1 Rank-1 and 85.7 mAP). Our method significantly closes gap between unsupervised and supervised performance on person re-identification, i.e., 92.2 vs. 94.1 top-1 accuracy in DukeMTMC to Market1501 transfer.

The main contributions of this paper can be summarized in four aspects: 
\begin{itemize}
\item We make the first attempt towards integrating clustering and feature learning in a unified framework through the label transfer method for UDA-ReID. It can online refine the predicted pseudo labels to improve the feature representation ability of the model on the target domain.
\item We propose a group-aware feature learning strategy based on label transfer to refine multi-group pseudo labels, which provides good latent pseudo-label groups for improving the quality of representation learning. 
\item The GLT framework achieves significant performance improvements compared to state-of-the-art approaches on Market$\to$Duke, Duke$\to$Market, Market$\to$MSMT, Duke$\to$MSMT ReID tasks. Even for the supervised learning methods, our algorithm is remarkably closing the gap.
\end{itemize}

\section{Related Work}
     \begin{figure*}[ht!]
        \centering
        \includegraphics[width=1.0\linewidth]{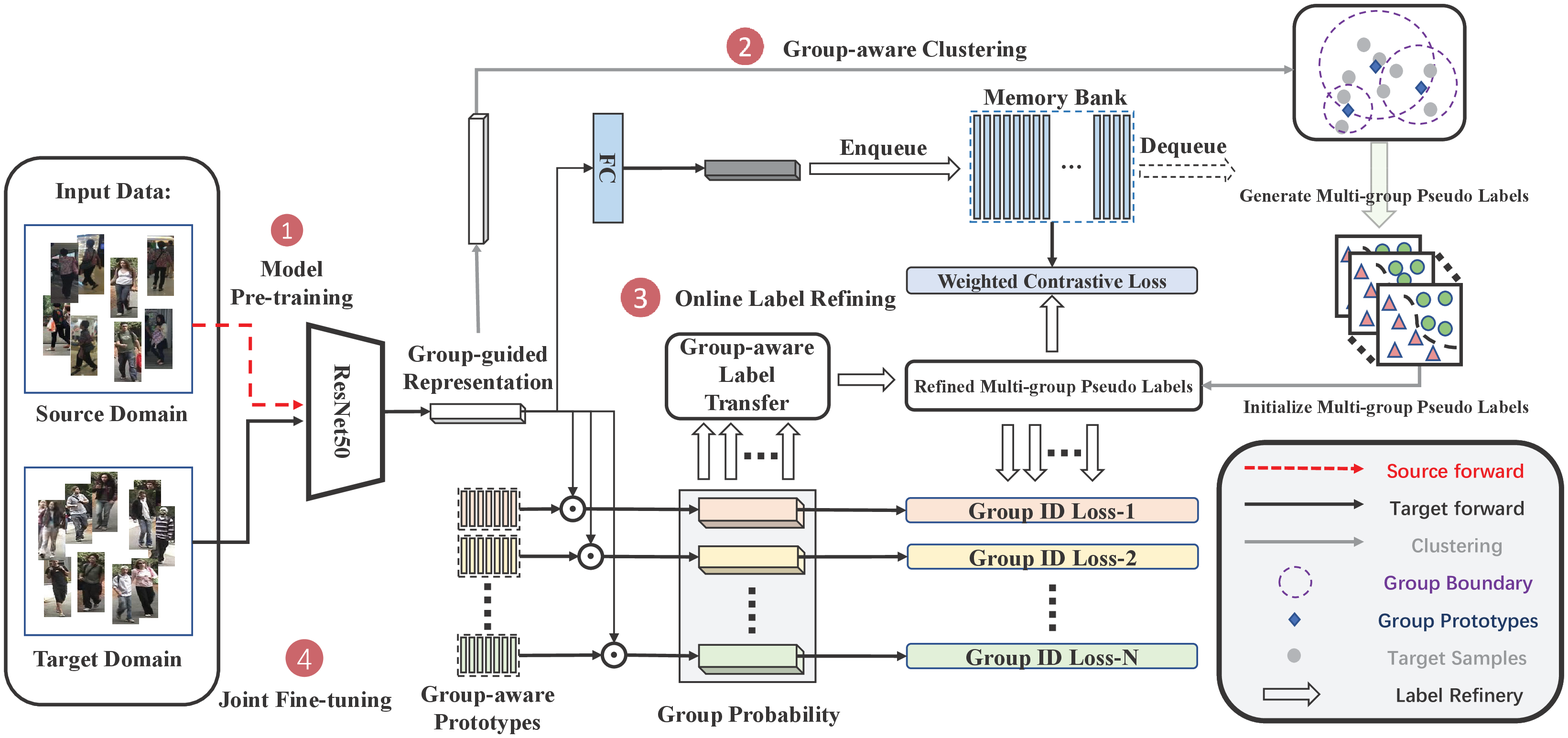}
        \caption{Illustration of our group-aware label transfer framework (GLT), where the group-aware label transfer and the ReID model are alternately optimized to improve each other towards the final objective of accurate person re-ID. In the model pre-training stage, we pre-train the network using source domain labeled data. In the group-aware clustering stage, we do clustering on the unlabeled target domain data using a multi-group strategy and assign multi-group pseudo labels based on the clustering results. In the label online refining stage, the label transfer algorithm treats the resulting label refining problem as an optimal transport problem, and refine the multi-group pseudo labels by linear programming. In the joint fine-tuning stage, we use the refined multi-group pseudo labels to train the networks. Stage 3 and Stage 4 are performed alternatively in the online scheme.}
        \label{fig:fm}
    \end{figure*}
    \noindent\textbf{Clustering-based Methods} in general generate hard or soft pseudo labels based on clustering results and then fine-tune/train the models based on the pseudo labels ~\cite{fan2018unsupervised,zhang2019self,yang2019selfsimilarity,ge2020mutual,yu2019unsupervised,zhong2019invariance, jin2020global,caron2020unsupervised,dai2020dual}. They are widely used due to their superior performance. PUL~\cite{fan2018unsupervised} iteratively obtains hard clustering labels and trains the model in a self-training manner. SSG~\cite{yang2019selfsimilarity} exploits the potential similarity for the global and local body parts, respectively, to build multiple independent clusters. These clusters are then assigned with labels to supervise the training. PAST~\cite{zhang2019self} optimizes the network with triplet-based loss and classification loss by appending a classification layer based on clustering results. 
	
	Pseudo label noise caused by unsupervised clustering is always an obstacle to the self-training. Such noisy labels would mislead the feature learning and impede the achievement of high performance. Recently, some methods introduce mutual learning among two/three collaborative networks to mutually exploit the refined soft pseudo labels of the peer networks as supervision \cite{ge2020mutual,zhai2020multiple}. To suppress the noises in the pseudo labels, NRMT \cite{zhao2020unsupervised} maintains two networks during training to perform collaborative clustering and mutual instance selection, which reduces the fitting to noisy instances by using mutual supervision and the reliable instance selection. These approaches need mutual learning of two or more networks and are somewhat complicated. Besides, the selection of reliable instances in NRMT~\cite{zhao2020unsupervised} is a hard selection, which requires a careful determination of threshold parameters and may lose the chance of exploiting useful information of the abandoned samples.

    \noindent\textbf{Domain translation based methods}~\cite{deng2018image,wei2018person} consider to translate target images by the source images. Then they use these translated source-to-target images and their corresponding ground-truth identities to fine-tune the target-domain model.
    PTGAN~\cite{wei2018person} utilizes the color consistency to impose pixel-level constraints during the domain translation.
    SPGAN~\cite{deng2018image} further minimizes the feature-level similarities between translated images and the original ones. This kind of approach does not apply to real-world scenarios, because the GAN-based method is hardly used to train and may introduce extra computing cost.
    
    \noindent\textbf{Memory Bank based methods} have been widely used for unsupervised representation learning which facilitates the introduction of contrastive loss~\cite{he2020momentum} for the general tasks. He \etal~\cite{he2020momentum} leverage the memory bank to better train the model to exploit the similarity between a sample and the instances in the global memory bank. Wang et al.~\cite{wang2020cross} propose using the memory bank to facilitate hard negative instance mining across batches. ECN~\cite{zhong2019invariance} leverages memory bank to enforce exemplar-invariance, camera-invariance, and neighborhood-invariance over global training data (instead of the local batch) for UDA person ReID. We introduce contrastive loss to the target instance memory bank to enable the joint optimization of positive pairs and negative pairs for a query/anchor sample over all the samples in the memory bank, which serves as our strong baseline.

    \section{Group-aware Label Transfer}
    We first introduce the overall architecture, then derive our group-aware label transfer framework and finally introduce the memory bank with weighted contrastive loss.

    \subsection{Clustering-based Method Revisit}
    \paragraph{Formulation.}
	Unsupervised domain adaptive person re-identification aims at adapting the model trained on a source domain dataset with annotations to another unlabeled target domain dataset. Formally, given the source domain data $\mathbb{D}_{s}=\left\{\left.\left(\boldsymbol{x}_{i}^{s}, \boldsymbol{y}_{i}^{s}\right)\right|_{i=1} ^{N_{s}}\right\}$ with the manual annotation, we can access target domain images $\mathbb{D}_{t}=\left\{\left.\boldsymbol{x}_{i}\right|_{i=1} ^{N_{t}}\right\}$ without any ground-truth labels.  $N_{s}$ and $N_{t}$ denote the number of the labeled source data and the unlabeled target data, respectively.
	
	\paragraph{Overview.} 
	Deep-cluster framework is the general pipeline due to the conciseness and effectiveness of this framework. To be specific, a ReID model $F(\cdot | \boldsymbol{\theta})$ is generally pre-trained via a supervised task on source domain data, where $\theta$ denotes the parameters of network. Then we adopt this pre-trained network to extract the features of target domain images which denote as $\left\{\boldsymbol{f_i}=F\left(\boldsymbol{x}_{i} | \boldsymbol{\theta}\right)\right\}|_{i=1} ^{N_{t}} \in \mathbb{R}^{D}$. 
	After that, the target domain images are grouped into $K$ classes by clustering these embedding features. Let $\{\tilde{\boldsymbol{y}}_{i}\}|^{N_t}_i\in\{1, \ldots, K\}$ denote the pseudo labels generated by target domain images $\{x_{i}\}|_{i}^{N_t}$, which contain the noisy label. 
	
	Then, to calculate a cross-entropy loss, a classification head $FC_k: \mathbb{R}^{D} \rightarrow \mathbb{R}^{K}$ is adopted to convert the embedding feature to a score vector, in the form of $p\left(\tilde{y}_{i} | \boldsymbol{x}_{i}\right)=\operatorname{softmax}\left(FC( F(\boldsymbol{x}_{i} | \boldsymbol{\theta}))\right)$.
	The network parameters $\theta$ and a learnable target-domain classifier $FC_k$ are then optimized with respect to a cross-entropy loss $\mathcal{L}_{i d}(\boldsymbol{\theta})$ and a triplet loss $\mathcal{L}_{t r i}(\boldsymbol{\theta})$ in the form of,
	\begin{equation}
	\begin{aligned} 
	\mathcal{L}_{i d}(\boldsymbol{\theta}) =-\frac{1}{m_{t}} \sum_{i=1}^{m_{t}} \log p\left(\tilde{y}_{i} | \boldsymbol{x}_{i}\right),
	\\ \mathcal{L}_{t r i}(\boldsymbol{\theta}) =\frac{1}{m_{t}} \sum_{i=1}^{m_{t}} \max \left(0,s^n_i-s^p_i+m\right), 
	\end{aligned}
	\label{eq:1}
	\end{equation}
	where $m_t$ is the mini-batch size of target data, $s^p_i$
	denotes the hardest positive cosine similarity, $s^n_i$
	denotes the hardest negative cosine similarity, scripts $i,p$ and $j,n$ indicate the indices of hardest positive and negative feature in each mini-batch for the sample $x_i$, and $m = 0$ denotes the triplet distance margin. 
	
	
	Such two steps, pseudo label generation by clustering and feature learning with pseudo labels, are alternated until the training converges. However, because of the domain gap, the pseudo-labels are not always reliable and there are noisy labels. This would mislead the feature representation learning towards learning noisy information. So, we need to adopt an online clustering strategy to refine those samples with wrong pseudo labels. As illustrated in Fig.~\ref{fig:fm}, the framework is alternated to gradually train the ReID model on the target domain with the guidance of pseudo labels.

    \subsection{Online Clustering by Group-aware Label Transfer}
    
    Each image $\boldsymbol x_i$ is transformed into $\boldsymbol f_i$ by a ReID model. We then compute a group probability $\boldsymbol p_i$ from this feature by mapping $\boldsymbol f_i$ to a set of $K$ trainable prototype vectors, $\boldsymbol C=\{c_1, c_2, \ldots c_K\}$. We denote the matrix by $\boldsymbol C$ whose columns are the $\{c_1, c_2, \ldots c_K\}$. Then, the pseudo labels $\boldsymbol Q \in [0,1]^{K \times N}$ are adopted to supervise the group probability $\boldsymbol P$, and we use Label Transfer to online update the pseudo labels as training goes on. We now first introduce how to update the prototypes online.

    \paragraph{Prototype Prediction Problem.} Our prototypes can be implemented by a non-parametric prototypical classifier or a parametric linear classifier. 
    The optimization can be viewed to optimize the cluster-assignment probability through the cross-entropy loss, where the prototypes $\boldsymbol C$ represent the trainable weights for a linear classifier. With k-means or DBSCAN clustering, the linear classifier has a fixed set of weights as the mean vectors for the representations in each cluster.  
    
    The prototypes are learned by the non-parametric or a parametric way noted above. Here, we will introduce the prototype prediction problem. Firstly, we rewrite the cross-entropy loss of eq.~\eqref{eq:1} by encoding the labels as posterior distributions $q\left(\tilde y_{i} | \boldsymbol{x}_{i}\right) \in \boldsymbol Q$, which is generated by clustering algorithm:
    \begin{equation}
    \begin{aligned}
    \mathcal{L}_{g}(p,q;\boldsymbol{\theta})= -\frac{1}{N} \sum_{N} q\left(\tilde y_{i} | \boldsymbol{x}_{i}\right) \log p\left(\tilde y_{i} | \boldsymbol{x}_{i}\right),
    \\ \text{where}\ p\left(\tilde y_{i} | \boldsymbol{x}_{i}\right) = \frac{exp(\frac{1}{\tau}\boldsymbol {f_i} \boldsymbol c_k)}{\sum_{k'} exp(\frac{1}{\tau}\boldsymbol {f_i} \boldsymbol c_{k'})},
    \label{eq:2}
    \end{aligned}
    \end{equation}
    where $\tau$ is a temperature parameter. If we set the posterior distributions $q\left(\tilde y_{i}| \boldsymbol{x}_{i}\right)=\delta\left(\tilde y_{i}-y_{i}\right)$ to be deterministic, this eq.~\eqref{eq:2} is similar to the general cross-entropy loss. We can use a non-parametric prototypical classifier performs or a parametric linear classifier to update the prototypes. This loss function is jointly minimized respect to the prototypes $\boldsymbol C$ and the image encoder parameters $\boldsymbol \theta$.
    
    \paragraph{Online Refining Pseudo Labels via Label Transfer.} On the Clustering-based UDA methods, inevitable noisy labels caused by the clustering procedure bring a negative effect on network training. To address this issue, we propose a label transfer method to correct the noisy pseudo labels in an online training scheme by combining the representation learning and clustering algorithm. 
    The general pipeline alternately tackles the optimization of $\boldsymbol q$ and representational learning $\boldsymbol p$. However, separate optimizations cannot well generate accuracy pseudo labels in representational learning because of restraining the interaction of these two steps. In order to make our method online, we try to utilize the prototypes and group probability to refine noisy pseudo labels.
    We compute refinery pseudo labels using the prototypes $\boldsymbol C$ such that all the examples in a batch are equally partitioned by the prototypes. This equipartition constraint ensures that the refinery pseudo labels for different images in several batches or a whole dataset are distinct, which preventing the trivial solution where every image has the same pseudo label:
    \begin{equation}
    \begin{aligned}
    \min_{p, q} \mathcal{L}_{g}(p,q;\boldsymbol{\theta}) \ \forall \tilde y_{i}: q\left(\tilde y_{i} | \boldsymbol{x}_{i}\right) \in\{0,1\} \\ \sum_{i=1}^{N} q\left(\tilde y_{i}=k | \boldsymbol{x}_{i}\right)=\frac{N}{k},
    \end{aligned}
    \label{eq:3}
    \end{equation}
    where each sample $x_i$ is assigned to exactly one label and the $N$ data points are split among the $K$ classes for alleviating a trivial solution.
    Inspired by~\cite{asano2019selflabelling}, they enforce an equal partition by constraining the matrix $\boldsymbol Q$ to belong to the transportable prototype. We also propose to adapt their solution to work on minibatches by restricting the transportable prototype to the minibatches or whole dataset.
    
    The objective in eq.~\eqref{eq:3} is combinatorial in $q$ and thus is very difficult to optimize by an end-to-end manner. However, the optimal transport algorithm~\cite{cuturi2013sinkhorn} is able to solve this issue. Let $P_{i,\tilde y_{i}}=p\left(\tilde y_{i} | \boldsymbol{x}_{i}\right) \frac{1}{N}$ be the ($i,\tilde y_{i}$) element of $K \times N$ matrix as the joint probabilities estimated by the model, which represents the group probability that the $i$-th sample is predicted to become the $\tilde y_{i}$-th pseudo-label. Similarly, $Q_{i,y_{i}}=q\left(\tilde y_{i} | \boldsymbol{x}_{i}\right) \frac{1}{N}$ is $K \times N$ matrix as the pseudo label matrix. Using the notation of~\cite{cuturi2013sinkhorn}, we relax matrix $\boldsymbol Q$ to be an element of the transportable prototype:
    \begin{equation}
    \begin{aligned}
    H(\boldsymbol r, \boldsymbol c):=\{\boldsymbol Q \in \mathbb{R}_{+}^{K \times N} | \boldsymbol Q \cdot \mathbf{1}=\boldsymbol w, \boldsymbol Q^{\top} \cdot \mathbf{1}= \boldsymbol c\}, \\\quad \boldsymbol c=\frac{1}{N} \cdot \mathbf{1}, \boldsymbol w=\frac{1}{N} \cdot \mathbf 1,
    \end{aligned}
    \label{eq:4}
    \end{equation}
    where $\mathbf{1}$ are vectors of all ones of the appropriate dimensions, so that $\boldsymbol c$ and $\boldsymbol w$ are the marginal projections of matrix $\boldsymbol Q$ onto its rows and columns, respectively. In our case, we require $\boldsymbol Q$ to be a matrix of conditional probability distributions that is similar to the number distribution of source domain.

    
    In practice, the computational complexity is enormous, including millions of data points and thousands of classes. Traditional algorithms for solving this problem are difficult to be applied. Thus, we adopt the Sinkhorn-Knopp algorithm~\cite{cuturi2013sinkhorn} to address this issue. This amounts to introducing a regularization term:
    \begin{equation}
    \min _{\boldsymbol Q \in H(\boldsymbol w, \boldsymbol c)}\langle \boldsymbol Q,-\log \boldsymbol P\rangle+\frac{1}{\lambda} \mathrm{H}\left(\boldsymbol Q\right),
    \label{eq:5}
    \end{equation}
    where $\langle.\rangle$ is the Frobenius dot-product between two matrices, $\boldsymbol R$ denotes to the number distribution of source domain, $\langle \boldsymbol Q,-\log \boldsymbol P\rangle= -\frac{1}{N} \sum_{i=1}^{N} \sum_{\tilde y_{i}=1}^{K} q\left(\tilde y_{i} | \boldsymbol{x}_{i}\right) \log p\left(\tilde y_{i} | \boldsymbol{x}_{i}\right)$, log is applied element-wise, $\mathrm H$ is the entropy function, $\mathrm H(Q) = -\sum_{q \in \boldsymbol{Q}} q \log q$ and $\lambda$ is a parameter that controls the smoothness of the mapping. We find that using a low $\lambda$ can have a strong entropy regularization, which generally leads to a trivial solution where all samples collapse into an unique representation and are all assigned uniformely to all prototypes. Hence, in practice we keep a high $\lambda$. According to the Sinkhorn-Knopp algorithm~\cite{asano2019selflabelling}, this minimizer of regularization term of eq.~\eqref{eq:5} can be written as:
    \begin{equation}
    \boldsymbol Q=\operatorname{diag}(\alpha) \boldsymbol P^{\lambda} \operatorname{diag}(\beta)=\operatorname{diag}(\alpha) (\boldsymbol C^T \boldsymbol F)^{\lambda} \operatorname{diag}(\beta),
    \label{eq:6}
    \end{equation}
    where $\boldsymbol F= [ \boldsymbol f_1, \ldots , \boldsymbol f_N]$ denotes to the feature vectors of images, $\boldsymbol C= [ \boldsymbol c_1, \ldots , \boldsymbol c_K]$ denotes to the prototypes of pseudo labels, $\alpha$ and $\beta$ are two renormalization vectors of scaling coefficients, the resulting matrix $\boldsymbol Q$ is also a probability matrix that represent the refining pseudo labels (see~\cite{cuturi2013sinkhorn} for a derivation). By the iterative Sinkhorn-Knopp algorithm~\cite{asano2019selflabelling}, the vectors $\alpha$ and $\beta$ can be obtained via a simple matrix scaling iteration:
    \begin{equation}\forall y: \alpha_{y} \leftarrow\left[\boldsymbol P^{\lambda} \beta\right]_{y}^{-1} \quad \forall i: \beta_{i} \leftarrow\left[\alpha^{\top} \boldsymbol P^{\lambda}\right]_{i}^{-1},
    \label{eq:7}
    \end{equation}
    where $\alpha_{0} = \boldsymbol c$ and $\beta_{0} = \boldsymbol w$. When $\lambda$ is very large, optimizing eq.~\eqref{eq:7} is mostly equivalent to optimizing eq.~\eqref{eq:6}. However, even for intermediate values of $lambda$, the results still tend to approximate the optimal solution~\cite{cuturi2013sinkhorn}. In the original optimal transport problem, $\lambda$ adjust the balance between the convergence speed and optimum approximation. In our case, we only regard the $\lambda$ as the hype-parameter, because we are more interested in obtaining a better performance in the final clustering and representation learning results, rather than solving the optimal transport problem.
    
    After using a group probability matrix $P$ to update the pseudo label matrix $Q$, we use the current $Q$ as the pseudo label to train the ReID model for optimizing the group probability $P$. Such two steps, pseudo label prediction and feature learning with pseudo labels are alternated until the training converges.

    \begin{algorithm}[tb]
    \caption{\small Group-aware Label Transfer} 
    \scriptsize
    \begin{algorithmic}[1]
    \REQUIRE Source domain data $\mathbb{D}_{s}=\left\{\left.\left(\boldsymbol{x}_{i}^{s}, \boldsymbol{y}_{i}^{s}\right)\right|_{i=1} ^{N_{s}}\right\}$, target domain images $\mathbb{D}_{t}=\left\{\left.\boldsymbol{x}_{i}\right|_{i=1} ^{N_{t}}\right\}$, multi-group $k$ values $\{k_1, \ldots, k_M\}$;
    
    \STATE Pre-training source-domain encoder $F(\cdot | \boldsymbol{\theta})$ by minimizing Eq.~\eqref{eq:1} on $\mathbb{D}_{s}$;
    \FOR{$k_m$ in $\{k_1, \ldots, k_M\}$}
    \STATE Generating multi-group pseudo labels $\boldsymbol Q_{k_m}\simeq\{\tilde{\boldsymbol{y}}_{i}\}|^{N_t}_i\in\{1, \ldots, k_m\}$ by clustering $\left\{F\left(\boldsymbol{x}_{i} | \boldsymbol{\theta}\right)\right\}|_{i=1} ^{N_{t}}$.;
    
    \FOR{$i$ in $[1, iterations]$}
    \STATE Updating parameters $\theta$ by the gradient descent of objective function equation~\eqref{eq:11} under the guidance of multi-group pseudo labels $\{Q_{k_m}\}|_{k=1}^{M}$;
    \FOR{$k_m$ in $\{k_1, \ldots, k_M\}$}
    \STATE Extracting the multi-group probability matrix $\boldsymbol P_{k_m}$ of target images $\mathbb{D}_{t}$ by the current model $F(\cdot | \boldsymbol{\theta})$;
    \WHILE{$||\boldsymbol \alpha_{j-1}-\boldsymbol \alpha_j||_1<0.1$}
    \STATE $\boldsymbol \alpha_{j} \leftarrow\left[\boldsymbol P_{k_m}^{\lambda} \boldsymbol \beta_{j-1}\right]^{-1}, \boldsymbol \beta_{j} \leftarrow\left[\boldsymbol \alpha_{j-1}^{\top} \boldsymbol P_{k_m}^{\lambda}\right]^{-1}$;
    \ENDWHILE
    \STATE $\boldsymbol Q'_{k_m}=\operatorname{diag}(\boldsymbol\alpha) \boldsymbol P_{k_m}^{\lambda} \operatorname{diag}(\boldsymbol\beta)$ \# Refining noisy multi-group pseudo labels;
    \ENDFOR
    
    Refining the multi-group pseudo labels $\boldsymbol Q_{k_m}=\boldsymbol Q'_{k_m}$;
    \ENDFOR
    \ENDFOR
    \end{algorithmic}
    \label{alg}
    \end{algorithm}

    \paragraph{Group-aware Label Transfer.} The actual identification number $k$ of the target domain is unknown, and it is hard to use the one value of $k$ to generate accurate pseudo labels. Therefore, group-based description~\cite{kim2020groupface} that involves common characteristics in the pseudo groups, can be useful to narrow down the set of candidates, which is beneficial to identify the exact persons. This group-aware strategy can cluster a person into multi-group clustering prototypes, which is able to efficiently embed a significant number of people and briefly describe an unknown person. Then, we introduce the group-aware strategy to assign implicit attribute group IDs to samples.
    
    In our formulation, we first adopt the multi-k setting $\{k_1, \ldots ,k_M\}$ to generate the multi-group pseudo label via multi-density DBSCAN or multi-k K-means. Then, we use them to train the ReID model:
    \begin{equation}
    \begin{aligned}
    \mathcal L_{G} = -\frac{1}{N} \sum_{m=1}^{M} \sum_{N} q_{k_{m}}\left(\tilde y_{i} | \boldsymbol{x}_{i}\right) \log p_{k_{m}}\left(\tilde y_{i} | \boldsymbol{x}_{i}\right),
    \label{eq:2}
    \end{aligned}
    \end{equation}
    
    Based on the multi-group pseudo labels, the group-aware label transfer adopts the multi-group optimization strategy to refine these multi-group pseudo labels. The different groups of pseudo label have different classification heads, while sharing the parameters of the feature extractor $\theta$ among them. In formulation, we optimize the multi-group objective functions of the type eq.~\eqref{eq:8}. 
    \begin{equation}
    \small
    \begin{aligned}
    \sum_{m=1}^{M} \min_{\boldsymbol Q_{k_m} \in H_{k_m}}(\langle \boldsymbol Q_{k_m},-\log \boldsymbol P_{k_m}\rangle+ \frac{1}{\lambda} \mathrm{H}(\boldsymbol Q_{k_m})), \\
    H_{k_m}:= \left\{\boldsymbol Q_{k_m} \in \mathbb{R}_{+}^{K_m \times N} | \boldsymbol Q_{k_m} \cdot \mathbf{1}=\boldsymbol r, \boldsymbol Q_{k_m}^{\top} \cdot \mathbf{1}=\boldsymbol c\right\}, 
    \\ \quad \boldsymbol c=\frac{1}{N} \cdot \mathbf{1}, \boldsymbol r=\frac{1}{N} \cdot \mathbf{1},
    \end{aligned}
    \label{eq:8}
    \end{equation}
    where $\{\boldsymbol P_{k_m} =\boldsymbol C_{k_m}^{T} \cdot \boldsymbol F\}|_{m=1}^{M}$ denotes to the multi-group probability and $\{\boldsymbol C_{k_m}\}|_{m=1}^{M}$ denotes to the multi-group prototypes. Each update involves $M$ matrix-vector multiplication with complexity $\mathcal{O}(MNK)$, so it is relatively quick.

    
	\subsection{Target Instance Memory Bank}
	We first describe our target instance memory bank module, with an updating mechanism. Then we show that a weighting scheme focusing on informative pairs is beneficial for contrastive loss with multiple positive samples.
	
	\paragraph{Updating mechanism.} We use the memory bank~\cite{he2019momentum} to memorize the features of the whole dataset. When the mini-batch of samples is coming, we maintain the memory bank as a queue of data samples. This allows us to reuse the feature embeddings from the immediate preceding mini-batches. The introduction of a queue decouples the memory bank size from the mini-batch size. Our memory bank size can be much larger than a typical mini-batch size and can be flexibly and independently set as a hyper-parameter. The samples in the dictionary are progressively replaced. The current mini-batch is enqueued to the dictionary, and the oldest mini-batch in the queue is removed. The dictionary always represents a sampled subset of all data, while the extra computation of maintaining this dictionary is manageable. Moreover, removing the oldest mini-batch can be beneficial, because its encoded keys are the most outdated and thus the least consistent with the newest ones.

	\paragraph{Weighted contrastive loss} Then we use the cosine distance to calculate the similarity $s$ of features between mini-batch and memory bank. Moreover, according to the pseudo labels, we split the samples to $L$ negative pairs and $K$ positive pairs, where weighted contrastive loss~\cite{sun2020circle} is suitable for this issue. Thus, weighted contrastive loss is proposed to better calculate the loss in terms of multiple positive samples, which is beneficial for the optimization of model:
	\begin{equation}
	\begin{aligned}
	\mathcal{L_{WCL}}=\log [1+  &\sum_{j=1}^{L}  \exp \left(\gamma \alpha^{n}_{j}\left(s^{n}_{j}-m\right)\right) \\
	&\sum_{i=1}^{K} \exp \left(-\gamma \alpha^{p}_{k}\left(s^{p}_{k}-1+m\right)\right) ],
	\end{aligned}
	\label{eq:9}
	\end{equation}
	where $\alpha^{n}_{j}=[m+s^n_j]_{+}$ and $\alpha^{p}_{k}=[1+m-s^p_k]_{+}$ are non-negative weighting factors respectively, $[\cdot]_{+}$ is the ``cut-off at zero'' operation, $m$ refers to a margin in order to better separate similarity, $\gamma$ is a scale factor, and $L$ and $K$ denote the number of negative pairs and positive pairs in the memory bank for the sample $x_i$ of mini-batch.

    We integrate the losses mentioned above. The total loss could be formulated as:
    \begin{equation}
    \mathcal L_{target} = \lambda_{tri} \mathcal L_{TRI} + \lambda_{g} \mathcal L_{G} + \lambda_{wcl} \mathcal{L}_{WCL}.
    \label{eq:11}
    \end{equation}
    where $\lambda_{tri}$, $\lambda_{g}$, and $\lambda_{wcl}$ are weighting factors. 

    \begin{table*}[ht!]
    \footnotesize
    \centering
    \caption{Performance comparison to the unsupervised domain adaptative person Re-ID state-of-the-art methods on DukeMTMC-reID~\cite{dukemtmc}, Market-1501~\cite{market} and MSMT17~\cite{wei2018person} datasets. ``BOT'' denotes to ``bag of tricks'' method~\cite{fastreid}, which is a strong baseline in the ReID task.}
    
    \begin{center}
      
        \begin{tabular}{P{5.0cm}|C{1cm}C{1.0cm}C{1.0cm}C{1.0cm}|C{1.0cm}C{1.0cm}C{1.0cm}C{1.0cm}}
            \hline
            \multicolumn{1}{c|}{\multirow{2}{*}{Methods}} & \multicolumn{4}{c|}{DukeMTMC$\to$Market1501} & \multicolumn{4}{c}{Market1501$\to$DukeMTMC} \\
            \cline{2-9}
            \multicolumn{1}{c|}{} & mAP & top-1 & top-5 & top-10 & mAP & top-1 & top-5 & top-10 \\ 
            \hline 
            UMDL~\cite{Peng2016Unsupervised} (CVPR'16) & 12.4&34.5& 52.6& 59.6& 7.3& 18.5& 31.4& 37.6 \\
            PTGAN~\cite{wei2018person} (CVPR'18)&- & 38.6 &-&66.1&- & 27.4 &-& 50.7 \\
            PUL~\cite{fan2018unsupervised} (TOMM'18) & 20.5 & 45.5 & 60.7 & 66.7 & 16.4 & 30.0 & 43.4 & 48.5 \\
            TJ-AIDL~\cite{wang2018transferable} (CVPR'18) & 26.5 & 58.2 & 74.8 & 81.1 & 23.0 & 44.3 & 59.6 & 65.0 \\
            SPGAN~\cite{deng2018image} (CVPR'18) & 22.8 & 51.5 & 70.1 & 76.8 & 22.3 & 41.1 & 56.6 & 63.0 \\
            ATNet~\cite{Liu2019cvpr}(CVPR'19) & 25.6& 55.7& 73.2& 79.4& 24.9 &45.1 &59.5& 64.2\\
            SPGAN+LMP~\cite{DengWeijian2018cvpr}(CVPR'18) &26.7 &57.7 &75.8 &82.4& 26.2&46.4 &62.3 &68.0 \\
            CFSM~\cite{chang2018disjoint} (AAAI'19) & 28.3 & 61.2 & - & - & 27.3 & 49.8 &- & -  \\
            BUC~\cite{lin2019aBottom} (AAAI'19) & 38.3 & 66.2 & 79.6 & 84.5 & 27.5 & 47.4 & 62.6 & 68.4 \\
            ECN~\cite{zhong2019invariance} (CVPR'19) & 43.0 & 75.1 & 87.6 & 91.6 & 40.4 & 63.3 & 75.8 & 80.4 \\
            UCDA~\cite{qi2019novel} (ICCV'19) & 30.9 & 60.4 & - & - & 31.0 & 47.7 & - & - \\
            PDA-Net~\cite{li2019cross} (ICCV'19) & 47.6 & 75.2 & 86.3 & 90.2 & 45.1 & 63.2 & 77.0 & 82.5 \\
            PCB-PAST~\cite{zhang2019self} (ICCV'19) & 54.6 & 78.4 & - & - & 54.3 & 72.4 & - & - \\
            SSG~\cite{yang2019selfsimilarity} (ICCV'19) & 58.3 & 80.0 & 90.0 & 92.4 & 53.4 & 73.0 & 80.6 & 83.2 \\
            ACT~\cite{Yang2019Asymmetric} (AAAI'20) & 60.6 & 80.5 & - & - & 54.5 & 72.4 & - & - \\
            MPLP+MMCL~\cite{WANG2020cvpr1} (CVPR'20) & 60.4 &84.4 &92.8& 95.0 & 51.4&72.4 &82.9& 85.0  \\
            DAAM~\cite{Huang2020aaai} (AAAI'20)& {48.8} & {71.3} & {82.4} & {86.3} & {53.1} & {77.8} & {89.9} & {93.7} \\
            AD-Cluster~\cite{zhai2020adcluster} (CVPR'20)& {68.3} & {86.7} & {94.4} & {96.5} & {54.1} & {72.6} & {82.5} & {85.5} \\ 
            MMT~\cite{ge2020mutual} (ICLR'20) & {71.2} & {87.7} & {94.9} & {96.9} & {65.1} & {78.0} & {88.8} & {92.5} \\
            NRMT~\cite{zhao2020unsupervised}(ECCV'20) & 71.7& 87.8& 94.6& 96.5& 62.2& 77.8& 86.9& 89.5 \\				
			B-SNR+GDS-H~\cite{jin2020global}(ECCV'20) & 72.5 & 89.3 & - & - & 59.7 & 76.7 &- &-\\				
			MEB-Net~\cite{zhai2020multiple}(ECCV'20) & {76.0} &{89.9}& {96.0}& {97.5}& {66.1}& {79.6} & 88.3& 92.2 \\
			\hline
            Source Pretrain&33.9&65.3&79.2&84.0 &35.2&53.3&67.7&73.1 \\
            Our GLT & \textbf{79.5} & \textbf{92.2} & \textbf{96.5} & \textbf{97.8} & \textbf{69.2} & \textbf{82.0} & \textbf{90.2} & \textbf{92.8} \\
            
            Supervised learning (BOT~\cite{fastreid})&85.7&94.1&-&- &75.8&86.2&-&-   \\
            \hline 
            
        \end{tabular}\\
        
        \begin{tabular}{P{5.0cm}|C{1.0cm}C{1.0cm}C{1.0cm}C{1.0cm}|C{1.0cm}C{1.0cm}C{1.0cm}C{1.0cm}}
            \hline
            \multicolumn{1}{c|}{\multirow{2}{*}{Methods}} & \multicolumn{4}{c|}{Marke1501$\to$MSMT17} & \multicolumn{4}{c}{DukeMTMC$\to$MSMT17} \\
            \cline{2-9}
            \multicolumn{1}{c|}{} & mAP & top-1 & top-5 & top-10 & mAP & top-1 & top-5 & top-10 \\ 
            \hline 
            PTGAN~\cite{wei2018person} (CVPR'18) & 2.9 & 10.2 & - & 24.4 & 3.3 & 11.8 & - & 27.4 \\    
            ENC~\cite{zhong2019invariance} (CVPR'19) & 8.5 & 25.3 & 36.3 & 42.1 & 10.2 & 30.2 & 41.5 & 46.8 \\
            SSG~\cite{yang2019selfsimilarity} (ICCV'19) & 13.2 & 31.6 &- & 49.6 & 13.3 & 32.2 & - & 51.2 \\
            DAAM~\cite{Huang2020aaai} (AAAI'20)& {20.8} & { 44.5} & {-} & {-} & { 21.6} & { 46.7} & {-} & {-} \\
            
            MMT~\cite{ge2020mutual} (ICLR'20) & 22.9 & 49.2 & {63.1} & {68.8} & 23.3 & 50.1 & 63.9 & {69.8} \\
            \hline

            Source Pretrain&7.3&19.8&29.4&34.5 &11.2&31.5&42.3&47.7 \\
            Our GLT & \textbf{26.5} & \textbf{56.6} & \textbf{67.5} & \textbf{72.0} & \textbf{27.7} & \textbf{59.5} & \textbf{70.1} & \textbf{74.2} \\
            Supervised learning (BOT~\cite{fastreid})&48.3&72.3&-&- &48.3&72.3&-&-   \\
            
            \hline

            \hline
            \end{tabular}
        \end{center}
        \label{tab:sota}
        \vspace{-10pt}
    \end{table*}
    
    \section{Experiments}

    \subsection{Datasets}
    We mainly evaluate our framework between three person ReID datasets, including DukeMTMC-reID \cite{dukemtmc}, Market-1501 \cite{market} and MSMT17 \cite{wei2018person}.
    DukeMTMC-reID \cite{dukemtmc} consists of 36,411 images of 702 identities for training and 702 identities for testing, where all images are captured from 8 cameras. 
    Market-1501 \cite{market} contains 12,936 images of 751 identities for training and 19,281 images of 750 identities for testing, which are captured by 6 cameras.
    MSMT~\cite{wei2018person} contains 126,441 images of 4101 identities, of which 1041 identities were used for training, and these images were captured by 15 cameras.
    We adopt Mean average precision (mAP) and CMC top-1/5/10 accuracy as evaluation.

    \subsection{Implementation Details}
    \label{sec:imp}
    \noindent\textbf{Training data organization.}
    For our joint training strategy, each mini-batch contains 64 source-domain images of 4 ground-truth identities (16 for each identity) and 64 target-domain images of 4 pseudo identities. The pseudo identities are assigned by the clustering algorithm and updated before each epoch. All images are resized to 256$\times$128,  and random perturbations are applied to each image, \eg randomly erasing, cropping and flipping.
    
    \noindent\textbf{Hyper-parameters.}
    We tune the hyper-parameters of the proposed framework on the task of Market$\to$Duke, and the chosen hyper-parameters are directly applied to all the other tasks.
    ADAM optimizer is adopted to optimize the networks with weighting factors $\lambda_{tri}=1$, $\lambda_{g}=1$, $\lambda_{mcl}=0.05$ and the triplet margin $m=0.3$. The initial learning rates ($lr$) are set to 0.00035 for person image encoders. The source-domain pre-training iterates for 30 epochs and the learning rate decreases to 1/10 of its previous value every 10 epochs. 
    We use the K-means or DBSCAN to initialize the pseudo label, and we utilize the group-aware label transfer to refine the pseudo label per epoch. We set $k=\{500,1000,1500,2000\}$ for all datasets in the K-means setting and set the $eps=\{0.56,0.58,0.60,0.62,0.64\}$ for all datasets in the DBSCAN setting. The results of DBSCAN are shown in the supplementary materials.

    \subsection{Comparison with the State-of-the-art Methods}
    We compare our proposed GLT with the state-of-the-art methods on four domain adaptation settings in Tab.~\ref{tab:sota}, \ie Duke$\to$Market, Market$\to$Duke, Market$\to$MSMT, and Duke$\to$MSMT. 
    Specifically, we surpass the clustering-based SSG~\cite{yang2019selfsimilarity} by considerable margins of $21.2\%$, $15.8\%$, $13.3\%$, and $14.4\%$ improvements in terms of mAP on these four tasks with simpler network architectures and no extra computation of local features. DAAM~\cite{Huang2020aaai} also adopts the clustering-based pipeline with the guidance of attention mechanism and our method shows a noticeable $30.7\%$, $15.1\%$, $5.7\%$ and $6.1\%$ improvements in terms of mAP on these four tasks without the help of attention mechanism. Meanwhile, our GLT adopts the one model to significantly surpass the two-model methods MMT~\cite{ge2020mutual} that uses the same backbone, showing a noticeable $8.3\%$, $4.1\%$, $3.6\%$ and $4.4\%$ improvements in terms of mAP. Our GLT significantly outperforms the second best UDA method MEB-Net by $3.5\% and 3.1\%$ in mAP accuracy, for Duke$\to$Market, Market$\to$Duke, respectively.
    Moreover, without any manual annotations, our unsupervised domain adaptive method significantly bridges the gap between unsupervised and supervised performance on person re-identification, i.e., $92.2\%$ vs. $94.1\%$ top-1 accuracy in Duke$\to$Market task~\cite{fastreid}.

    \subsection{Ablation Studies}
    \label{sec:abla}
    
    In this section, we evaluate each components of our proposed framework by conducting ablation studies on Duke$\to$Market and Market$\to$Duke tasks. The experimental results are shown in Tab.~\ref{tab:ablation1}.
    
    \noindent\textbf{Effectiveness of proposed components.}
    We conduct extensive ablation studies to investigate the effectiveness of the proposed components in Tab.~\ref{tab:ablation1}, i.e., label transfer, group-aware label transfer, and target instance memory bank. 
    When introducing the label transfer algorithm, the performance is improved by $7.8\%$ rank-1 accuracy and $6.9\%$ mAP on Duke$\to$Market task. By sensibly adopting memory bank with weighted contrastive loss, better representation learning enables the model to increase the rank-1 accuracy from $88.1\%$ to $89.0\%$ and mAP from $66.4\%$ to $68.2\%$, respectively. 
    By taking account of the group-aware label transfer, the performance is improved by $11.3\%$ rank-1 accuracy and $3.0\%$ mAP, which indicates that the group-aware label transfer can provides good latent pseudo-label groups for improving the quality of representation learning. In addition, group-aware strategy without the guidance of label transfer bring a little improvement due to the lack of label transfer for pseudo label refinery. This illustrates the importance of group-aware label transfer module, not just group-aware strategy.
    These experimental results prove the necessity and effectiveness of our group-aware label transfer for correcting noise pseudo labels.

    \begin{table}[ht!]
        
        \footnotesize
        \caption{Ablation studies for our proposed framework on individual components. \textbf{TMB}: target instance memory bank. \textbf{G}: baseline with the multi-group pseudo labels. \textbf{LT}: label transfer. \textbf{GLT}: group-aware label transfer.}
        \centering
        
        \label{tab:ablation1}
        \begin{center}
            
            \begin{tabular}{P{3.2cm}|C{0.7cm}C{0.7cm}C{0.7cm}C{1.0cm}}
                \hline
                \hline
                \multicolumn{1}{c|}{\multirow{2}{*}{Methods}}& \multicolumn{4}{c}{Duke$\to$Market}  \\
                \cline{2-5}
                \multicolumn{1}{c|}{} & mAP & top-1 & top-5 & top-10  \\ 
                \hline 
                Supervised learning&85.7&94.1&-&-   \\
                \hline
                Source Pretrain &38.5&65.7&79.2&84.0   \\
                Baseline &59.5& 80.3& 90.4 &93.0 \\ 
                Baseline + G &63.0& 84.7& 91.4 &93.7 \\
                \hline
                Self-labeling~\cite{asano2019selflabelling} &40.9& 66.6& 80.6 &86.3 \\ 
                \hline
                Baseline + TMB & {62.4} & {82.1} & {91.8} & {94.9} \\
                Baseline + LT& {66.4} & {88.1} & {94.4} & {96.2} \\
                Baseline + TMB + LT & {68.2} & {89.0} & {94.5} & {96.1}  \\
                \hline
                Baseline + G + TMB & {65.2} & {84.3} & {94.8} & {96.7}  \\
                Baseline + G + GLT & {78.7} & {91.4} & {96.7} & {97.4}  \\
                Baseline + G + TMB + GLT  & \textbf{79.5} & \textbf{92.2} & \textbf{96.5} & \textbf{97.8}  \\
                \hline
                \hline
            \end{tabular}
        \end{center}
        \vspace{-10pt}
    \end{table}

    \begin{table}[ht!]
        \small
        \centering
         \begin{tabular}{C{3.3cm}|C{0.5cm}C{0.9cm}|C{0.5cm}C{0.9cm}}
            \hline
            \hline
            \multicolumn{1}{c|}{\multirow{2}{*}{Loss Functions}} & \multicolumn{2}{c|}{Duke$\to$Market} & \multicolumn{2}{c}{Market$\to$Duke} \\
            \cline{2-5}
            \multicolumn{1}{c|}{} & mAP & top-1 & mAP & top-1 \\ 
            \hline 
            \hline
            \\[-2.3ex]
            GLT w/o TRI loss & 77.3 & 90.9 & 67.1 & 81.2 \\
            GLT w/o WCL loss & 77.9 & 90.8 & 66.4 & 80.7 \\
            \hline
            GLT & \textbf{79.5} & \textbf{92.2}  & \textbf{69.2} & \textbf{82.0} \\
            \hline
            \hline
        \end{tabular}
        \caption{Evaluation of the effectiveness of triplet loss and weighted contrastive loss, ``w/o WCL'' denotes to using the weighted contrastive loss~\cite{he2019momentum}.} 
        \label{fig:rp1}
        
        \vspace{-10pt}
    \end{table}

    \noindent\textbf{Effectiveness of loss functions.}
    We validate the effectiveness of multi-group ID loss, weighted contrastive loss and triplet loss in our proposed GLT framework. Triplet loss brings $3.2\%$ and $2.1\%$ improvements in terms of mAP, and weighted contrastive loss brings $2.6\%$ and $2.8\%$ improvements in terms of mAP in the Market$\to$Duke and Duke$\to$Market tasks, respectively. Triplet loss utilizes the hardest information within a mini-batch of instance pairs in each iteration. Meanwhile, Weighted contrastive loss in the memory bank mines the global negative and positive instances among the samples of whole dataset. The both losses can help each other.
    
    
    \section{Conclusion}
    To solve the problem of UDA person ReID, we propose a novel framework, the group-aware label transfer method (GLT), to combine the accurate pseudo-label prediction and effective ReID representation learning in one unified optimization objective. The holistic and immediate interaction between these two steps in the training process can greatly help the UDA person ReID task. To achieve this goal, we consider the online label refinement problem as an optimal transport problem, which explores the minimal cost of assigning M samples to N pseudo labels.
    More importantly, we introduce a group-aware strategy to assign implicit attribute group IDs to samples. The combination of the online label refining algorithm and the group-aware strategy can better correct the noisy pseudo label in an online fashion and narrow down the search space of the target identity. Our method not only achieves state-of-the-art performance but also significantly closes the gap between supervised and unsupervised performance on person re-identification.

    

    \section{Acknowledgement}
    This work was supported by the National Key R$\&$D Program of China under Grand 2020AAA0105702, National Natural Science Foundation of China (NSFC) under Grants U19B2038.
    
    {\small\bibliographystyle{plain}
    \bibliography{references}}
    
\end{document}